\documentclass[10pt,twocolumn,letterpaper]{article}
\pdfoutput=1
\usepackage{wacv}
\usepackage{times}
\usepackage{epsfig}
\usepackage{graphicx}
\usepackage{amsmath}
\usepackage{amssymb}
\usepackage{float}
\usepackage{authblk}
\usepackage{bm}
\makeatletter
\renewcommand\AB@affilsepx{ , \protect\Affilfont}
\makeatother
\def\wacvPaperID{571} 

\wacvfinalcopy 

\ifwacvfinal
\def\assignedStartPage{9876} 
\fi
\ifwacvfinal
\usepackage[breaklinks=true,bookmarks=false]{hyperref}
\else
\usepackage[pagebackref=true,breaklinks=true,colorlinks,bookmarks=false]{hyperref}
\fi

\ifwacvfinal
\else
\fi

\begin{document}

\title{Temporal-Aware Self-Supervised Learning for 3D Hand Pose and Mesh Estimation in Videos}

\author[1]{Liangjian Chen}
\author[2]{Shih-Yao Lin}
\newcommand\CoAuthorMark{\footnotemark[\arabic{footnote}]} 
\author[3]{Yusheng Xie\protect\CoAuthorMark\thanks{Work done outside of Amazon.}}
\author[4]{Yen-Yu Lin}
\author[1]{Xiaohui Xie}
\affil[1]{University of California, Irvine}
\affil[2]{Tencent America}
\affil[3]{Amazon}
\affil[4]{National Chiao Tung University}
\affil[ ]{\textit {\{liangjc2,xhx\}@ics.uci.edu}}
\affil[ ]{\textit {mike.lin@ieee.org}}
\affil[ ]{\textit {yushx@amazon.com}}
\affil[ ]{\textit {lin@cs.nctu.edu.tw }}

\maketitle
\vspace*{-1.2cm}
\begin{abstract}

Estimating 3D hand pose directly from RGB images is challenging but has gained steady progress recently by training deep models with annotated 3D poses. 
However annotating 3D poses is difficult and as such only a few 3D hand pose datasets are available, all with limited sample sizes. 
In this study, we propose a new framework of training 3D pose estimation models from RGB images without using explicit 3D annotations, i.e., trained with only 2D information. 
Our framework is motivated by two observations: 1) Videos provide richer information for estimating 3D poses as opposed to static images; 
2) Estimated 3D poses ought to be consistent whether the videos are viewed in the forward order or reverse order. 
%
We leverage these two observations to develop a self-supervised learning model called temporal-aware self-supervised network (TASSN).
By enforcing temporal consistency constraints, TASSN learns 3D hand poses and meshes from videos with only 2D keypoint position annotations. 
%
Experiments show that our model achieves surprisingly good results, with 3D estimation accuracy on par with the state-of-the-art models trained with 3D annotations, highlighting the benefit of the temporal consistency in constraining 3D prediction models. 

%
%
%
\end{abstract}

\begin{figure}[t]
\centering
\includegraphics[width=0.48\textwidth]{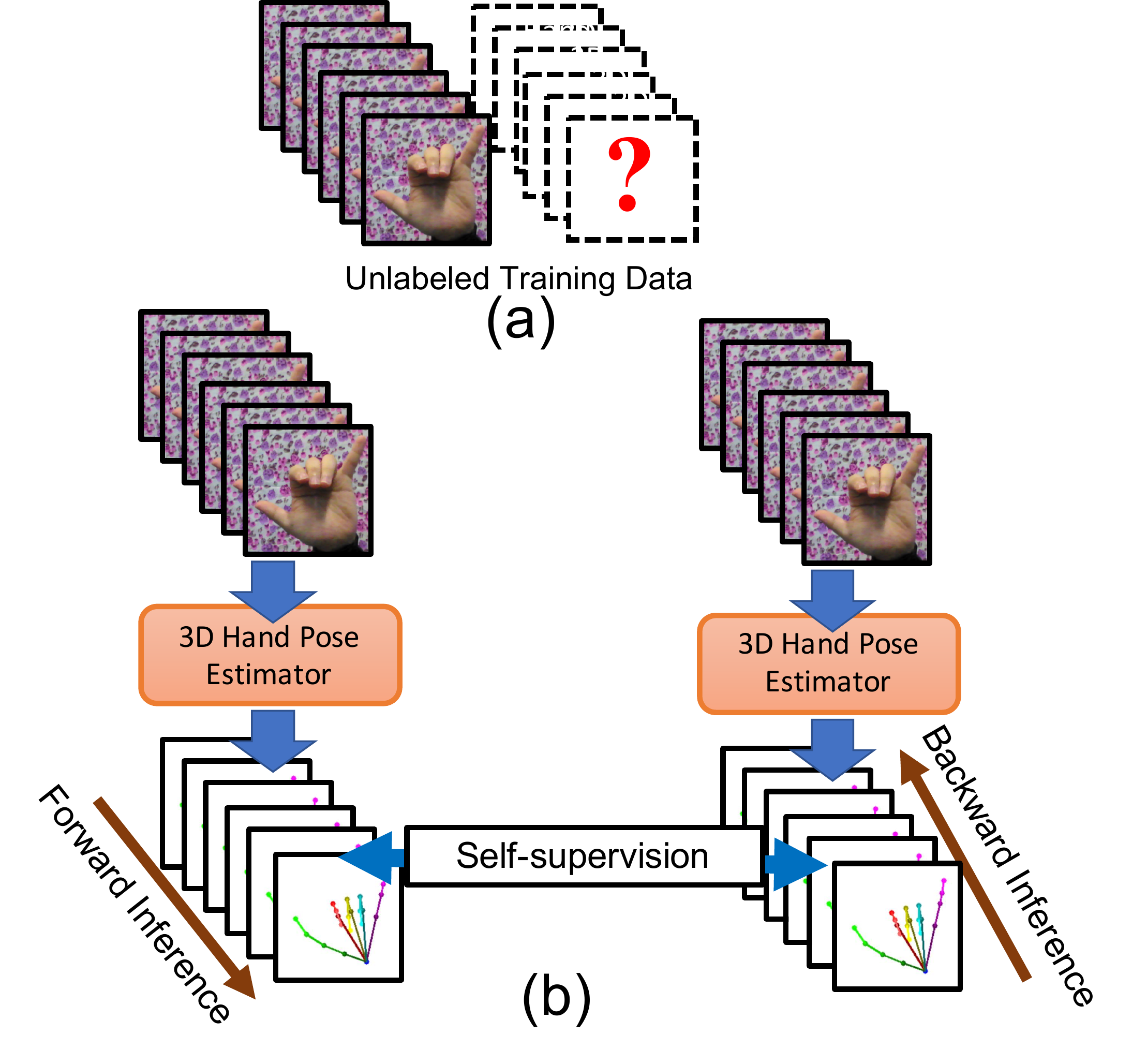}
\vspace{-0.5cm}
\caption{Motivation and idea: (a) Training a robust 3D hand pose estimator from RGB images relies on plenty images with 3D hand pose annotations, but obtaining 3D annotations from 2D images is quite difficult; (b) We leverage bi-directional temporal consistency in videos and enable hand pose estimators to make more plausible predictions. It turns out that the hand pose estimator can be derived in a self-supervised fashion without using 3D annotations.}
\label{fig:coreidea}
\end{figure}

\vspace{-0.5cm}
\section{Introduction}
\label{sec:intro}
3D hand estimation is an important research topic in computer vision due to a wide range of potential applications, such as sign language translation~\cite{zafrulla2011american}, robotics~\cite{antoshchuk2018gesture}, movement disorder detection and monitoring, and human-computer interaction (HCI)~\cite{lin2013airtouch, hung2016re,mikeicpr10}. 

Depth sensors and RGB cameras are popular devices for collecting hand data.
However, depth sensors are not as widely available as RGB cameras and are much more expensive, which has limited the applicability of hand pose estimation methods developed upon depth images.  
Recent research interests have shifted toward estimating 3D hand poses directly from RGB images by utilizing color, texture, and shape information contained in RGB images. 
%
%
Some methods carried out 3D hand pose estimation from monocular RGB images~\cite{cai2018weakly,iqbal2018hand,zb2017hand}. 
More recently, progresses have been made on estimating 3D hand shape and mesh from RGB images~\cite{baek2019pushing,boukhayma20193d,ge2019handshapepose,zhang2019end,mm-hand, chen_wacv21_dataset,kong2020sia,kong2019adaptive,zhao2020image,zhao2020topk,zhao2020improved}. 
Compared to poses, hand meshes provide richer information required by many immersive VR and AR applications.
%
Despite the advances, 3D hand pose estimation remains a challenging problem due to the lack of accurate, large-scale 3D pose annotations. 

In this work, we develop a new approach to 3D hand pose and mesh estimation by taking the following two observations into account.
First, most existing methods rely on training data with 3D information, but capturing 3D information from 2D images is intrinsically difficult. 
Although there are a few datasets providing annotated 3D hand joints, the amount is too small to train a robust hand pose estimator. 
Second, most studies focus on hand pose estimation from a single image.
Nevertheless, important applications based on 3D hand poses, such as augmented reality (AR), virtual reality (VR), and sign language recognition, are usually carried out in videos.
%

According to the two observations, our approach exploits video temporal consistency to address the uncertainty caused by the lack of 3D joint annotations on training data.
Specifically, our approach, called {\em temporal-aware self-supervised network (TASSN)}, can learn and infer 3D hand poses without using annotated 3D training data.
Figure~\ref{fig:coreidea} shows the motivation and core idea of the proposed TASSN.
%
TASSN explores video information by embedding a temporal structure to extract spatial-temporal features.
%
We design a novel temporal self-consistency loss, which helps training the hand pose estimator without requiring annotated 3D training data. 
%
%
In addition to poses, we estimate hand meshes since meshes bring salient evidences for pose inference. 
%
With meshes, we can infer silhouettes to further regularize our model.
%
%
%
%
The main contributions of this work are given below:
\begin{enumerate}
\item We develop a temporal consistency loss and a reversed temporal information technique for extracting spatio-temporal features. 
To the best of our knowledge, this work makes the first attempt to estimate 3D hand poses and meshes without using 3D annotations.
\item An end-to-end trainable framework, named temporal-aware self-supervised networks (TASSN), is proposed to learn an estimator without using annotated 3D training data.
The learned estimator can jointly infer the 3D hand poses and meshes from video.
%
\item Our model achieves high accuracy with 3D prediction performance on par with state-of-the-art models trained with 3D ground truth.

\end{enumerate}

\section{Related Work}
\label{sec:related}

%
%
%


\subsection{3D Hand Pose Estimation from Depth Images}
Since depth images contain surface geometry information of hands, they are widely used for hand pose estimation in the literature~\cite{wan2018dense,Yuan_2018_CVPR,deng2017hand3d,Wu18HandPose,ge2018hand,Ge_2018_ECCV,li2018point,Chen2018Generating,chen2018generating_arxiv}.
Most existing work adopts regression to fit the parameters of a deformed hand model~\cite{makris2015model,joseph2016fits,khamis2015learning,wan2018dense}. 
Recent work \cite{ge2018hand,Ge_2018_ECCV} extracts depth image features and regress the joints through PointNet \cite{Qi_2017_CVPR}. 
Wu~\etal~\cite{Wu18HandPose} leverage the depth image as the intermediate guidance and conduct an end-to-end training framework.
Despite the effectiveness, the aforementioned methods highly rely on accurate depth maps, and are less practical in the daily life since depth sensors are not available in many cases due to the high cost.
 
\subsection{3D Hand Pose Estimation from RGB Images}

Owing to the wide accessibility of RGB cameras, estimating 3D hand poses from monocular images becomes an active research topic~\cite{cai2018weakly,iqbal2018hand,mueller2018ganerated,tekin2019h+,yang2018disentangling,zb2017hand} and significant improvement has been witnessed.
These methods use convolutional neural networks (CNN) to extract features from RGB images.
Zimmermann and Brox \cite{zb2017hand} feed these features to the 3D lift network and camera parameter estimation network for depth regression. 
Building on Zimmermann and Brox's work, Iqbal~\etal \cite{iqbal2018hand} add depth maps as intermediate guidance while Cai~\etal~\cite{cai2018weakly} propose a weakly supervised approach to utilize depth maps for regularization.
However, these methods suffer from limited training data since 3D hand annotations are hard to acquired. 
Also, they all dismiss the temporal information.

\subsection{3D Hand Mesh Estimation}

3D hand mesh estimation is an active research topic~\cite{ge2019handshapepose,boukhayma20193d,baek2019pushing,joo2018total,zhang2019end}. 
%
%
Methods in~\cite{boukhayma20193d,baek2019pushing,zhang2019end} estimate hand meshes by using a pre-defined hand model, named MANO~\cite{romero2017embodied}. 
Due to the high degree of freedom of hand gestures, hand meshes lie in a high dimensional space. 
The MANO model serves as a kinematic and shape prior of meshes and can help reduce the dimension.
However, since MANO is a linear model, it is not able to capture the nonlinear transformation for hand meshes~\cite{ge2019handshapepose}. 
Thus, mesh estimators based on MANO suffer from this issue.
On the other hand, Ge~\etal~\cite{ge2019handshapepose} regress 3D mesh vertices through graphical convolutional neural network (GCN) with down-sampling and up-sampling.
Their work achieves the state-of-the-art performance, but it is trained on a dataset with 3D mesh ground truth which is even more difficult to label than 3D joint annotations. %
This drawback limits its applicability in practice.

\begin{figure*}[t]
\centering 
\hspace{-1cm}
\includegraphics[width=0.93\textwidth]{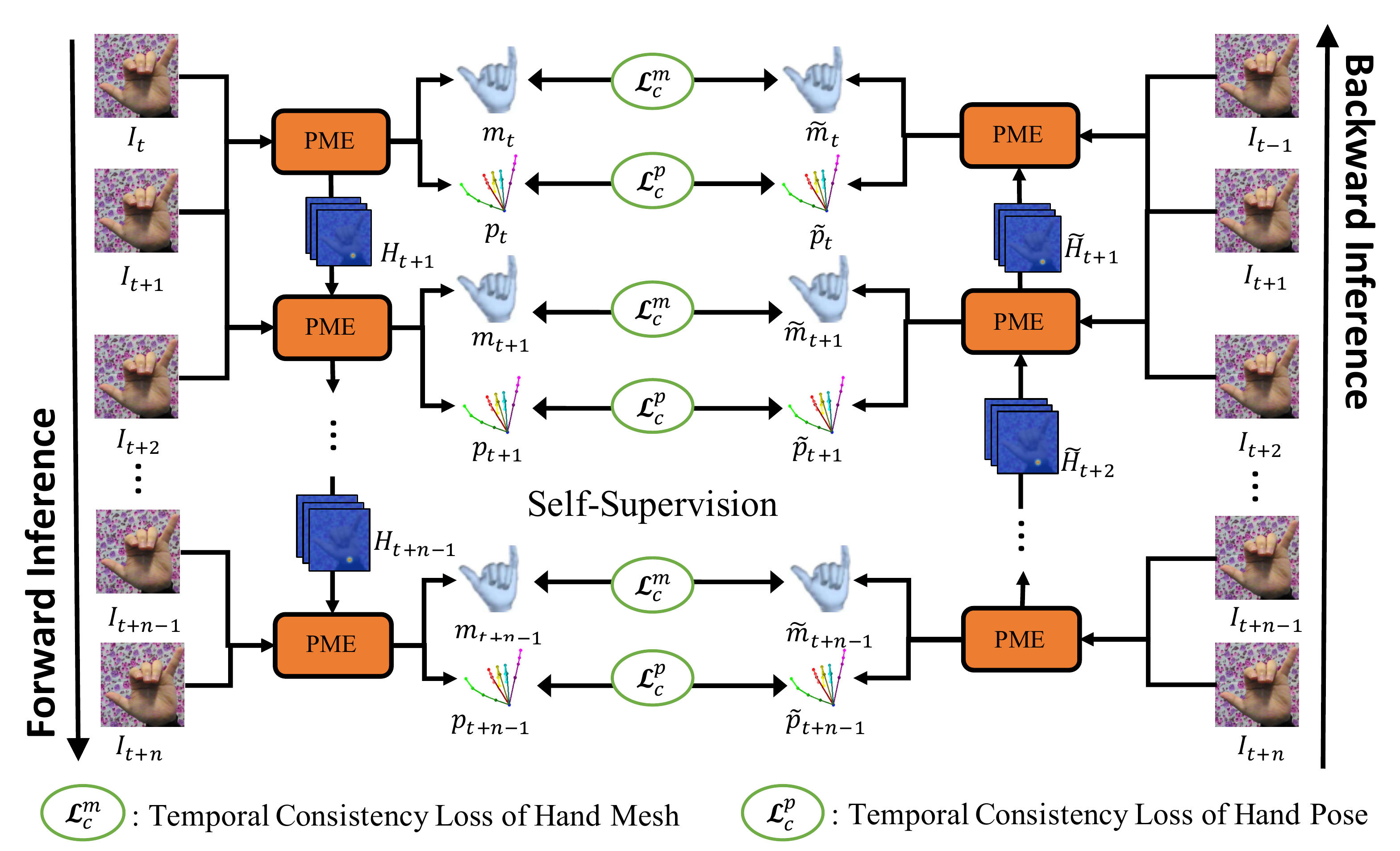}
\caption{Overview of the proposed TASSN. 
TASSN involves both forward and backward inference to utilize temporal information.
Namely, the hand poses estimated by forward and backward inference should be consistent. 
Our hand pose estimator leverages this observation and can be trained by using self-supervised learning without the need of 3D hand joint labels.
Moreover, with the constraints of temporal consistency, either forward or backward inference can gain more accurate hand pose estimation results.  
%
%
%
%
%
%
%
%
}
\label{fig:2}
\end{figure*}

\subsection{Self-supervised Learning}
Self-supervised learning~\cite{doersch2015unsupervised,pathak2017learning,Debidatta2019temporal} is a type of training methodologies, where training data are automatically labeled by exploiting existing information within the data.
With this training scheme, manual annotations are not required for a given training set.
This scheme is especially beneficial when data labeling is difficult or the data size is exceedingly large.
Self-supervised learning has been applied to hand pose estimation.
Similar to ours, the method in~\cite{Debidatta2019temporal} adopts temporal cycle consistency for self-supervised learning. 
However, this method uses soft nearest neighbors to solve the video alignment problem, which is not applicable to 3D pose and mesh estimation.
Simon~\etal~\cite{simon2017hand} adopt multi-view supervisory signals to regress 3D hand joint locations. 
While their approach resolves the hand self-occlusion issue using multi-view images, it in the training stage requires 3D joint annotations, which are difficult and expensive to get in this task. 
Another attempt of using self-supervised learning for hand pose estimation is presented in~\cite{wan2019self}, where an approach leveraging a massive amount of unlabeled depth images is proposed. 
However, this approach may be limited due to the high variations of depth maps in diverse poses, scales, and sensing devices.
Instead of leveraging multi-view consistency or depth consistency, the proposed self-supervised scheme relies on temporal consistency, which is inexpensive to get and does not require 3D keypoint annotations.

\begin{figure*}[t]
\centering 
\includegraphics[width=0.95\textwidth]{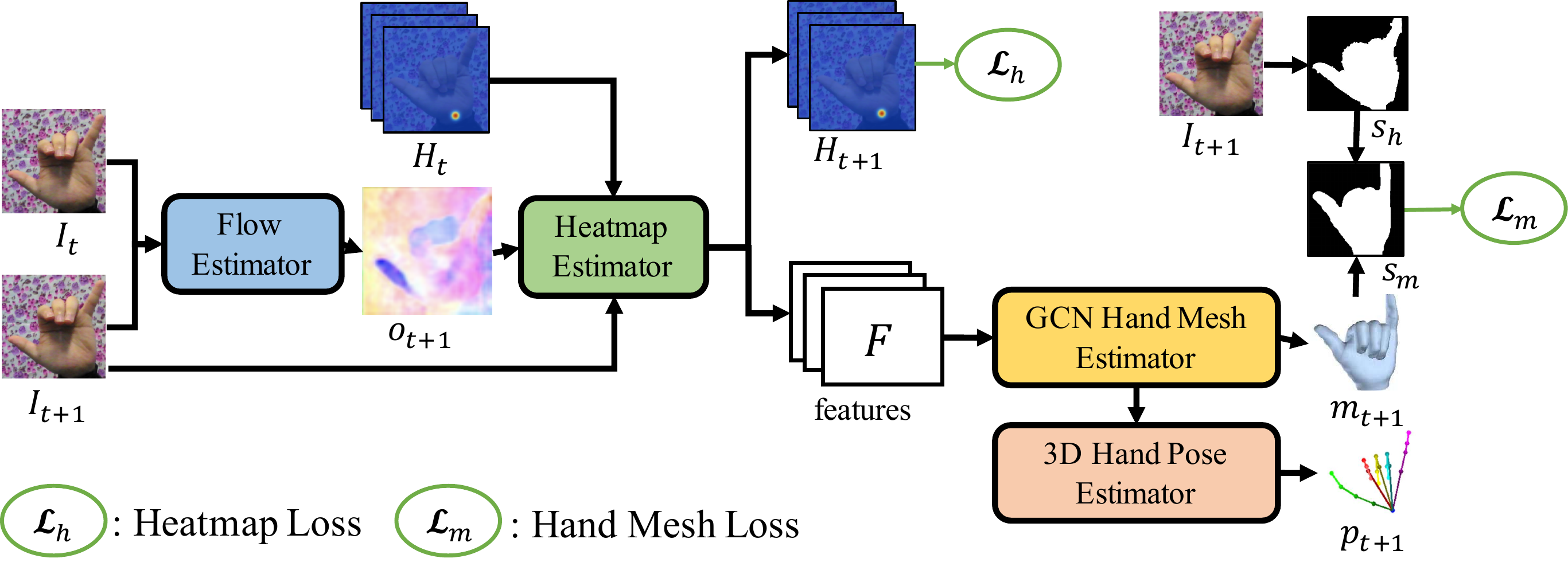}\label{fig:architecture}
\caption{Network architecture of the pose and mesh estimation (PME) module. 
PME module consists of four sub-modules, including the flow, 2D keypoint heatmap, 3D hand mesh, and 3D hand pose estimators. 
The flow estimator computes the optical flow $\bm{o}_{t+1}$ from two consecutive frames $\bm{I}_t$ and $\bm{I}_{t+1}$. 
%
%
With $\bm{I}_{t+1}$, $\bm{o}_{t+1}$, and $\bm{H}_{t}$, the 2D heatmap estimator computes the keypoint heatmap $\bm{H}_{t+1}$ at timestamp $t+1$, as well as extract the image features. 
Based on the extracted image features, the 3D hand pose and mesh estimators predict the 3D hand pose $\bm{p}_{t+1}$ and mesh $\bm{m}_{t+1}$ at timestamp $t+1$. 
Two loss terms, the heatmap loss $\bm{\mathcal{L}}_{h}$ and the hand mesh loss $\bm{\mathcal{L}}_{m}$, are used for optimization.}
\label{fig:pme}
\vspace{-0.5cm}
\end{figure*}


\section{Proposed Method}
\label{sec:method}
We aim to train a 3D hand pose estimator from videos without 3D hand joint labels.
%
%
%
To tackle the absence of 3D annotations, we adopt the temporal information from hand motion videos, and address the ambiguity caused by the lack of 3D joint ground truth.
Specifically, we present a novel deep neural network, named temporal-aware self-supervised networks (TASSN).
By developing the temporal consistency loss on the estimated hand gestures in a video, TASSN can learn and infer 3D hand poses through self-supervised learning without using any $3$D annotations.

\subsection{Overview}
\label{sec:overview}

Given an RGB hand motion video $\bm{x}$ with $N$ frames, $\bm{x}=\{\bm{I}_{1},...,\bm{I}_N\}$, we aim at estimating $3$D hand poses in this video, where $\bm{I}_t\in\mathbb{R}^{3\times W\times H}$ is the $t$-th frame, and $W$ and $H$ are the frame width and height, respectively.
The $3$D hand pose at frame $t$, $\bm{p}_{t}\in \mathbb{R}^{3 \times K}$, is represented by a set of $K$ $3$D keypoint coordinates of the hand.
%
Figure~\ref{fig:2} illustrates the network architecture of TASSN.

%
Leveraging the temporal consistency properties of videos, the hand poses and meshes predicted in the forward and backward inference orders can perform mutual supervision. 
%
Our model can be fine-tuned on any target dataset using this self-supervised learning and the temporal consistency is a good substitute for the hard-to-get 3D ground truth. 
TASSN alleviates the burden of annotating 3D ground-truth of a dataset without significantly sacrificing model performance. 

Recent studies~\cite{ge2019handshapepose,zhang2019end} show that training pose estimators with hand meshes improves the performance because hand meshes can act as intermediate guidance for hand pose prediction.
To this end, we propose a hand pose and mesh estimation (PME) module, which jointly estimates the 2D hand keypoint heatmaps, 3D hand poses and meshes from every two adjacent frames $\bm{I}_i$ and $\bm{I}_{i+1}$.

\subsection{Pose and Mesh Estimation Module}
\label{sec:pem}

The proposed PME module consists of four estimator sub-modules, including flow estimator, 2D keypoint heatmap estimator, 3D hand mesh estimator, and 3D hand pose estimator.
Given two consecutive frames as input, it estimates the 3D hand pose and mesh.
Figure~\ref{fig:pme} shows its network architecture.

\vspace{-0.1in}
{\flushleft {\bf Flow Estimator}}:
To capture temporal clues from a hand gesture video, we adopt FlowNet~\cite{ilg2017flownet} to estimate the optical flow   $\bm{o}_{t+1} \in \mathbb{R}^{2\times W\times H}$ between two consecutive frames $\bm{I}_{t}$ and $\bm{I}_{t+1}$.
In forward inference, FlowNet computes $\bm{o}_{t+1}$, the motion from frame $\bm{I}_{t}$ to frame $\bm{I}_{t+1}$.
In backward inference, FlowNet computes the reverse motion.

\vspace{-0.1in}
{\flushleft {\bf Heatmap Estimator}}:
Our heatmap estimator computes 2D hand keypoints and generates the features for the 3D hand pose and mesh estimators.
The estimated 2D keypoint heatmaps are denoted by $\bm{H} \in \mathbb{R}^{K\times W \times H}$, where $K$ represents the number of keypoints.
We adopt a two stacked hourglass network~\cite{newell2016stacked} to infer the hand keypoint heatmaps $\bm{H}$ and compute the features $\bm{F}$.
%
We concatenate $\bm{I}_{t+1}$, $\bm{o}_{t+1}$, and $\bm{H}_t$ as input to the stacked hourglass network, which produces heatmaps $\bm{H}_{t+1}$, as shown in Figure~\ref{fig:pme}.
The estimated $\bm{H}_{t+1}$ includes $K$ heatmaps $\{\bm{H}_{t+1}^k\in \mathbb{R}^{W \times H}\}_{k=1}^K$, where $\bm{H}_{t+1}^k$ expresses the confidence map of the location of the $k$th keypoint.
The ground truth heatmap $\bm{\bar{H}}^k_{t+1}$ is the Gaussian blur of the Dirac-$\delta$ distribution centered at the ground truth location of the $k$th keypoint.
The heatmap loss $\bm{\mathcal{L}}_{h}$ at frame $t$ is defined by 
\begin{equation}
\label{eq:heatmap}
    \bm{\mathcal{L}}_{h} =\frac{1}{K} \sum_{k = 1}^K||\bm{H}^k_t - \bm{\bar{H}}_t^k||_F^2.
\end{equation}


{\flushleft {\bf 3D Hand Mesh Estimator}}:
Our 3D hand mesh estimator is developed based on Chebyshev spectral graph convolution network (GCN)~\cite{ge2019handshapepose}, and it takes hand features $\bm{F}$ as input and infers the 3D hand mesh.
The output hand mesh $\bm{m}_{t}\in \mathbb{R}^{3 \times C}$ is represented by a set of $3$D mesh vertices, where $C$ is the number of vertices in a hand mesh.

To model hand mesh, we use an undirected graph $\bm{G}(\bm{V},\bm{E})$, where $\bm{V}$ and $\bm{E}$ are the vertex and edge sets, respectively.
The edge set $\bm{E}$ can be represented by an adjacent matrix $\bm{A}$, where $\bm{A}_{i,j} = 1$ if edge $e(i,j) \in \bm{E}$, otherwise $\bm{A}_{i,j} = 0$.
The normalized Laplacian normal matrix of $\bm{G}$ is obtained via $\bm{L} = \bm{I} - \bm{D}^{-\frac{1}{2}}\bm{A}\bm{D}^{-\frac{1}{2}}$, where $\bm{D}$ is the degree matrix and $\bm{I}$ is the identity matrix.
%
%
Since $\bm{L}$ is a positive semi-definite matrix~\cite{bruna2013spectral}, it can be decomposed as $\bm{L} = \bm{U\Lambda U}^T$, where $\bm{\Lambda} = diag(\lambda_1, \lambda_2,... , \lambda_C)$, and $C$ is the number of vertices in $\bm{G}$.

We follow the setting in~\cite{defferrard2016convolutional}, and set the convolution kernel to $\bm{\hat{\Lambda}} = diag(\sum_{i=0}^S\alpha_i\lambda_1^i, ... ,\sum_{i=0}^S\alpha_i\lambda_C^i)$, where $\alpha$ is the kernel parameter. 
The convolutional operations in $\bm{G}$ can be calculated by $\bm{F'} = \bm{U\hat{\Lambda} U}^T\bm{F\theta}_i = \sum_{i=0}^{S} \alpha_i\bm{L}^i\bm{F\theta}_i$,
where $\bm{F}\in \mathbb{R}^{N\times F_{\text{in}}}$ and $\bm{F}' \in \mathbb{R}^{N\times F_\text{out}}$ indicate the input and output features respectively, $S$ is a preset hyperparameter used to control the receptive field, and $\bm{\theta}_i \in \mathbb{R}^{F_\text{in} \times F_\text{out}}$ is trainable parameter set used to control the number of output channels.

%
The Chebyshev polynomial is used to reduce the model complexity by approximating convolution operations, leading to the output features $\bm{F'} = \sum_{i=0}^{S} \alpha_iT_i(\hat{\bm{L}})\bm{\theta}_i$ where $T_k(x)$ is the $k$-th Chebyshev polynomial and $\hat{\bm{L}} = 2\bm{L} / \lambda_{max}- \bm{I}$ is used to normalize the input features.

We adopt the scheme in~\cite{defferrard2016convolutional,ge2019handshapepose} to construct the hand mesh in a coarse-to-fine manner. 
We use the multi-level clustering algorithm for coarsening the graph, and then store the graph at each level and the mapping between graph nodes in every two consecutive levels.
In forward inference, the GCN first up-samples the node features according to the stored mappings and graphs and then preforms the graph convolutional operations.
%


\vspace{-0.1in}
{\flushleft {\bf Mesh Silhouette Constraint}}:
In our model, without 3D mesh ground truth, the model tends to collapse to any kind of mesh as long as it is temporally consistent. 
To avoid this issue, we introduce the mesh loss $\bm{\mathcal{L}}_{m}$ to calculate the difference between the silhouette of the predicted hand mesh $\bm{s}_{t}$ and the ground-truth silhouette $\bm{\bar{s}_t}$ at frame $t$. The silhouette loss is defined by
\begin{equation}
\label{eq:silhouette}
    \bm{\mathcal{L}}_{m} = ||\bm{s}_t - \bm{\bar{s}}_t||_F^2.
\end{equation} 

\begin{figure}[t]
\begin{center}
	\begin{tabular}{cccccc}
		\includegraphics[width=0.065\textwidth]{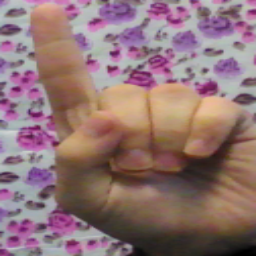}
		\includegraphics[width=0.065\textwidth]{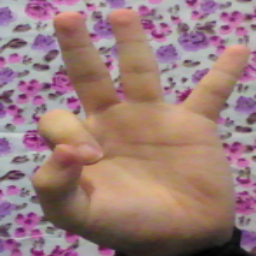}
		\includegraphics[width=0.065\textwidth]{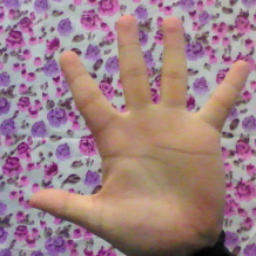}
		\includegraphics[width=0.065\textwidth]{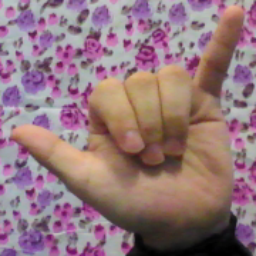}
		\includegraphics[width=0.065\textwidth]{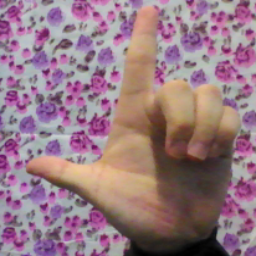}
		\includegraphics[width=0.065\textwidth]{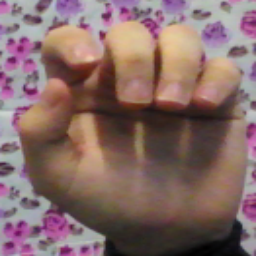}
	\end{tabular}
	\begin{tabular}{ccccc}
		\includegraphics[width=0.065\textwidth]{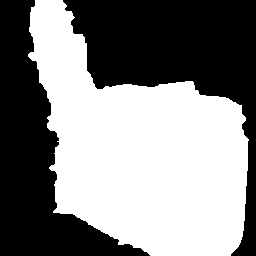}
		\includegraphics[width=0.065\textwidth]{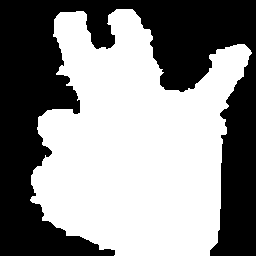}
		\includegraphics[width=0.065\textwidth]{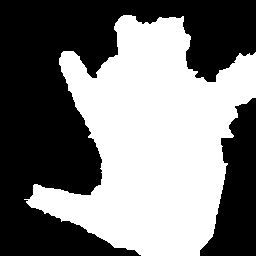}
		\includegraphics[width=0.065\textwidth]{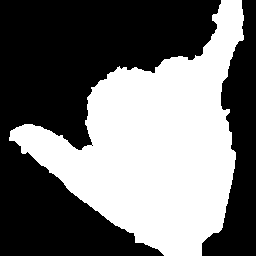}
		\includegraphics[width=0.065\textwidth]{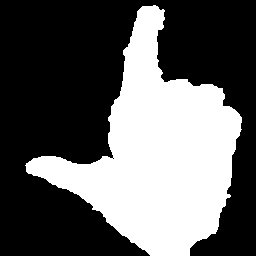}
		\includegraphics[width=0.065\textwidth]{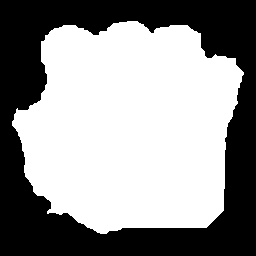}
	\end{tabular}
	\begin{tabular}{ccccc}
    	\includegraphics[width=0.065\textwidth]{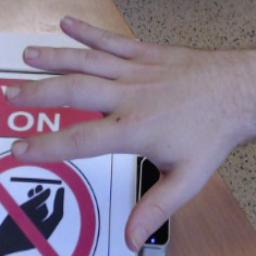}
    	\includegraphics[width=0.065\textwidth]{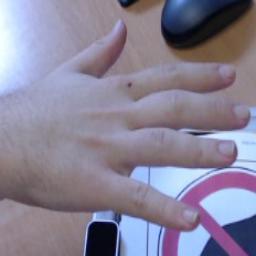}
    	\includegraphics[width=0.065\textwidth]{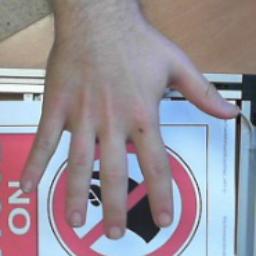}
    	\includegraphics[width=0.065\textwidth]{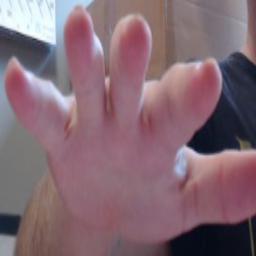}
    	\includegraphics[width=0.065\textwidth]{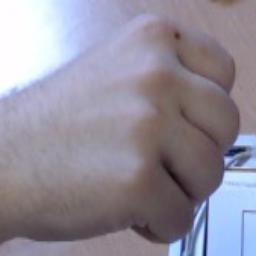}
    	\includegraphics[width=0.065\textwidth]{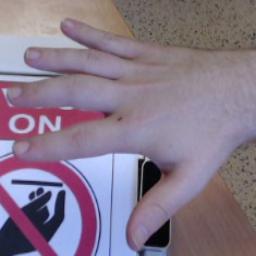}
    \end{tabular}
    \begin{tabular}{ccccc}
        \includegraphics[width=0.065\textwidth]{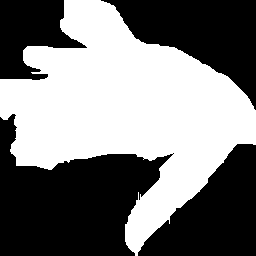}
    	\includegraphics[width=0.065\textwidth]{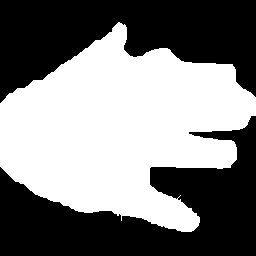}
    	\includegraphics[width=0.065\textwidth]{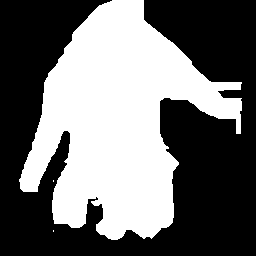}
    	\includegraphics[width=0.065\textwidth]{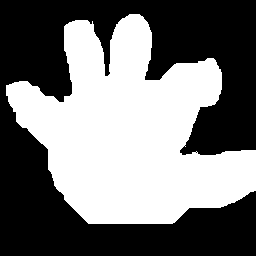}
    	\includegraphics[width=0.065\textwidth]{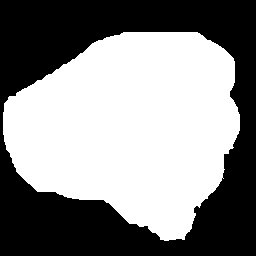}
    	\includegraphics[width=0.065\textwidth]{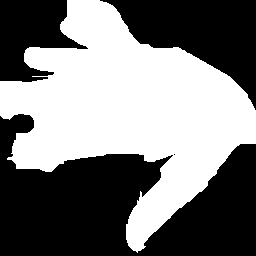}
    \end{tabular}
\end{center}
\vspace{-0.1in}
\caption{Examples of our estimated silhouettes.
The first and third rows show the training images in STB and MHP datasets, respectively.
The second and the fourth rows show the estimated silhouettes by our method.
}
\label{fig:grabcut}
\end{figure}

To obtain $\bm{\bar{s}}_t$, we use GrabCut~\cite{rother2004grabcut} to estimate the hand silhouettes from the training images.
Some silhouettes estimated from training images are shown in Figure~\ref{fig:grabcut}.
The silhouette of our predicted hand mesh $\bm{s}_{t}$ is obtained by using the neural rendering approach in~\cite{kato2018neural}.


{\flushleft {\bf 3D Hand Pose Estimator}}:
The proposed 3D pose estimator directly infers 3D hand keypoints $\bm{p}_t$ from the predicted hand mesh $\bm{m}_t$.
Taking the mesh as the input, we adopt a network of two stacked GCNs, which has a similar structure to that used in 3D hand mesh estimator.
We add a pooling layer to each GCN to extract the pose features from the mesh. 
Those pose features are then fed to two fully connected layers to regress the 3D hand pose $\bm{p}_t$.



\subsection{Temporal Consistency Loss}

Due to the lack of 3D keypoint annotations, conventional supervised learning schemes no longer work in model training.
We propose a temporal consistency loss $\bm{\mathcal{L}}_{c}$ to solve this problem.
Figure~\ref{fig:2} shows the idea of our approach.
Given a video clip with $n$ frames, we feed every two adjacent frames $\{\bm{I}_{i}, \bm{I}_{i+1}\}^{t+n}_{i=t}$ to PME module for hand mesh and pose estimation, \ie, $\{\bm{p}_i$, $\bm{m}_i\}^{t+n}_{i=t}$.
TASSN analyzes the temporal information according to their relative input orders. 
Thus, we can reverse the input order from $\{\bm{I}_{i}, \bm{I}_{i+1}\}$ to $\{\bm{I}_{i+1}, \bm{I}_{i}\}$ to infer the pose and mesh in $\bm{I}_{i}$ from $\bm{I}_{i+1}$. 
With this reversed temporal measurement (RTM) technique, we can infer the hand pose and mesh from the reversed temporal order. 
%
%
We denote the estimated pose and mesh in the reversed order as $\{\bm{\tilde{p}}_{i}$, $\bm{\tilde{m}}_{i}\}^{t+n}_{i=t}$. 
%
As shown in Figure~\ref{fig:2}, the prediction results estimated by the PME module in both forward and backward inference must be consistent with each other since the same mesh and pose are estimated at any frame.
The temporal consistency loss on hand pose $\bm{\mathcal{L}}_{c}^p$ and mesh $\bm{\mathcal{L}}_{c}^m$ can be computed by
\begin{equation}
    \bm{\mathcal{L}}_{c}^p =\frac{1}{n} \sum_{i = t}^{t+n}||\bm{{p}}_{i}\ - \bm{\tilde{p}}_{i}||_F^2, 
\end{equation}
\begin{equation}
\label{eq:temporal_mesh}
    \bm{\mathcal{L}}_{c}^m =\frac{1}{n} \sum_{i = t}^{t+n}||\bm{{m}}_{i}\ - \bm{\tilde{m}}_{i}||_F^2.
\end{equation}
The temporal consistency loss $\bm{\mathcal{L}}_{c}$ is defined as the summation of
$\bm{\mathcal{L}}_{c}^m$ and $\bm{\mathcal{L}}_{c}^p$, \ie,
\begin{equation}
\label{eq:temporal}
    \bm{\mathcal{L}}_{c} = \lambda^m\bm{\mathcal{L}}_{c}^m + \lambda^p\bm{\mathcal{L}}_{c}^p,
\end{equation}
where $\lambda^m$ and $\lambda^p$ are the weights of the corresponding losses.

\subsection{TASSN Training}
\label{sec:tassn_training}
Suppose we are given an unlabeled hand pose dataset $\bm{X}$ for training, which contains $M$ hand gesture videos, $\bm{X}=\{\bm{x}^{(i)}\}_{i=1}^M$, where video $\bm{x}^{(i)}=\{I_{1},...,I_N\}$ consists of $N$ frames. 
We divide each training video into several video clips. 
Each training video clip $\bm{v}$ is with $n$ frames, \ie, $\bm{v}= \{I_{t}, I_{t+1},...,I_{t+n}\}$. 
With the losses defined in Eq.~(\ref{eq:heatmap}), Eq.~(\ref{eq:silhouette}), and Eq.~(\ref{eq:temporal}), the objective for training the proposed TASSN is
\begin{equation}
    \bm{\mathcal{L}}=\lambda_s\bm{\mathcal{L}}_{m}+\lambda_h\bm{\mathcal{L}}_{h}+\bm{\mathcal{L}}_{c},
\end{equation}
where $\lambda^s$ and $\lambda^h$ denote the weights of the loss $\bm{\mathcal{L}}_{m}$ and the loss $\bm{\mathcal{L}}_{h}$, respectively. 
The details of parameter setting are given in the experiments.

\begin{figure}[t]
\begin{center}
	\begin{tabular}{ccccc}
		\includegraphics[width=0.085\textwidth]{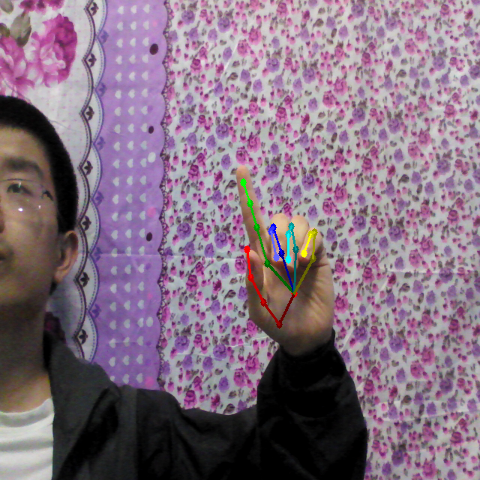}
		\includegraphics[width=0.085\textwidth]{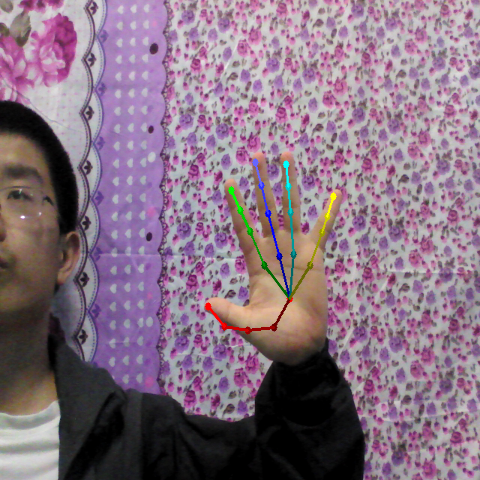}
		\includegraphics[width=0.085\textwidth]{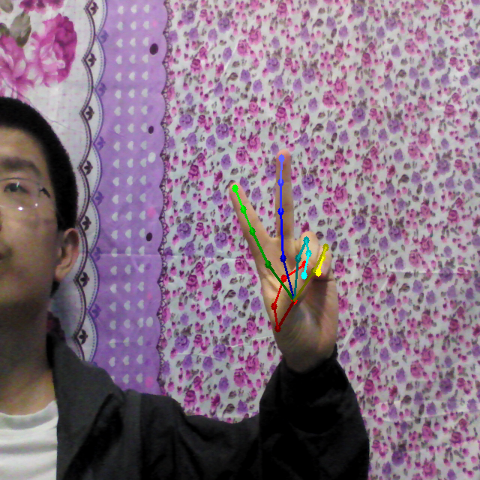}
		\includegraphics[width=0.085\textwidth]{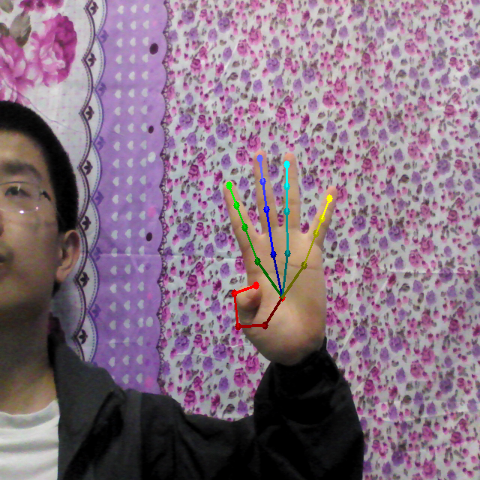}
		\includegraphics[width=0.085\textwidth]{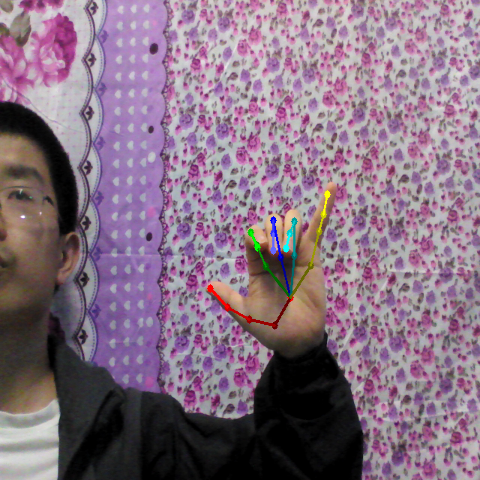}
	\end{tabular}
	\begin{tabular}{ccccc}
		\includegraphics[width=0.085\textwidth]{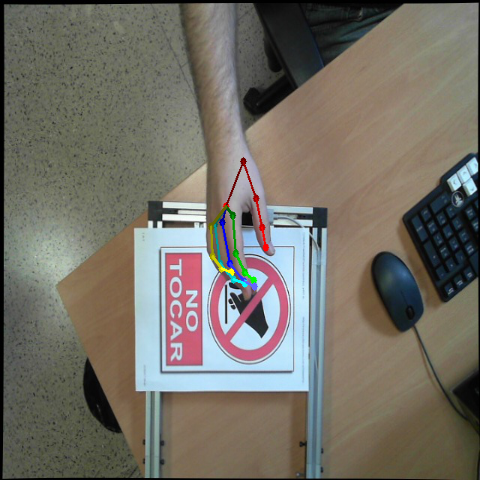}
		\includegraphics[width=0.085\textwidth]{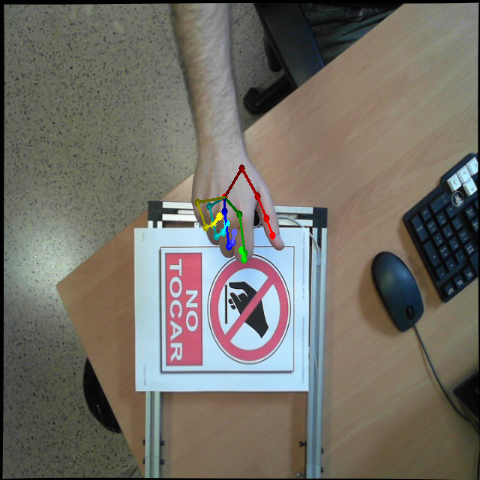}
		\includegraphics[width=0.085\textwidth]{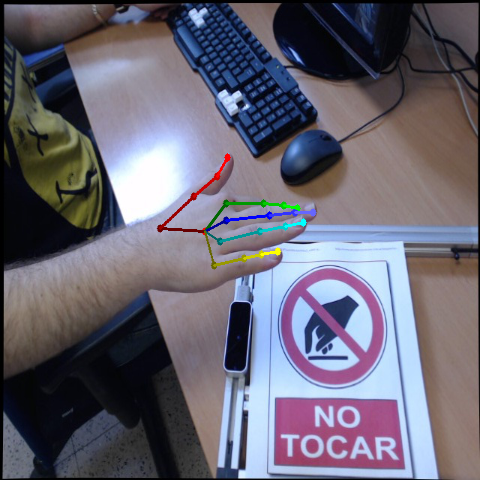}
		\includegraphics[width=0.085\textwidth]{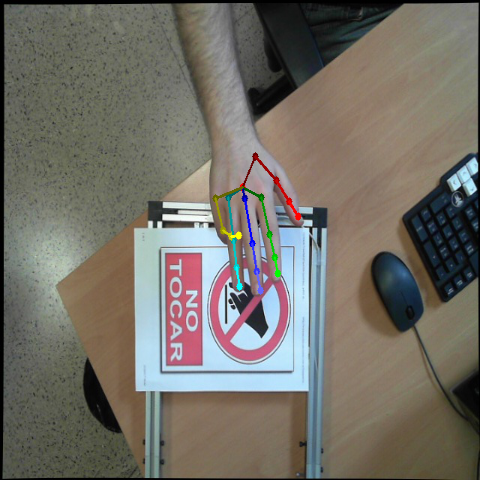}
		\includegraphics[width=0.085\textwidth]{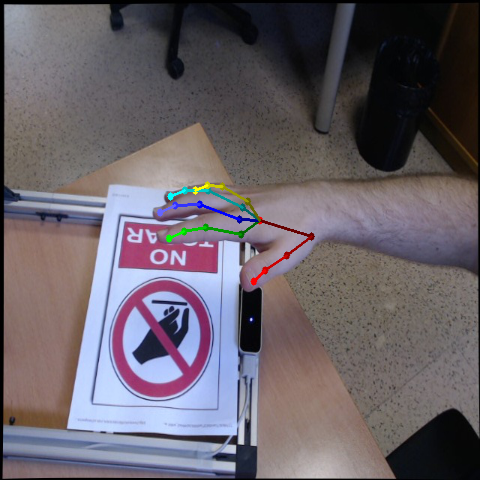}
	\end{tabular}
\end{center}
\vspace{-0.3cm}
\caption{Some examples of the two hand pose datasets used for evaluation.
The first row shows examples of the STB dataset~\cite{zhang20163d} and the second row gives examples of the MHP dataset~\cite{gomez2017large}. Both datasets include real hand video sequences performed by different subjects and have 3D hand keypoint annotations 
}
\label{fig:dataset}
\end{figure}

\section{Experiments Setting}
\label{sec:setting}

\subsection{Datasets for Evaluation}
We evaluate our approach on two hand pose datasets, Stereo Tracking Benchmark Dataset (STB)~\cite{zhang20163d} and Multi-view 3D Hand Pose dataset (MHP)~\cite{gomez2017large}.
These two datasets include real hand video sequences performed by different subjects and 3D hand keypoint annotations are provided for the hand video sequences.

For the STB dataset, we adopt its SK subset for training and evaluation.
This subset contains $6$ hand videos, each of which has $1,500$ frames.
%
%
Following the train-validation split setting used in \cite{ge2019handshapepose}, we use the first hand video as the validation set and the rest videos for training.

The MHP dataset includes $21$ hand motion videos.
Each video provides hand color images and different kinds of annotations for each sample, including the bounding box and the 2D and 3D location on the hand keypoints.
%

%
The following scheme of data pre-processing is applied to both STB and MHP datasets.
We crop the hand from the original image by using the center of hand and the scale of the hand.
Thus, the center of the hand is located at the center of the cropped images, and the cropped image covers the whole hand.
We then resize the cropped image to $256\times256$. 
As mentioned in~\cite{cai2018weakly,zb2017hand}, the STB and MHP datasets use the palm center as the center of the hand.
We use the mechanism introduced by~\cite{cai2018weakly} to change the center of hand from the palm center to the joint of wrist.

\subsection{Metric}
We follow the setting adopted in previous work~\cite{zb2017hand,ge2019handshapepose} and use {\em average End-Point-Error} (EPE) and  {\em Area Under the Curve} (AUC) on the {\em Percentage of Correct Keypoints} (PCK) between threshold $20$ millimeter (mm) and $50$mm ($\text{AUC}_\text{20-50}$) as the two metrics. 
Beside, we adopt AUC on PCK between threshold $0$mm and $50$mm ($\text{AUC}_\text{0-50}$) as the third metrics for evaluating 3D hand pose estimation performance. 
The measuring unit of EPE is millimeter (mm).

\subsection{Implementation Details}
\label{sec:traing}

We implement our TASSN by using PyTorch.
In training phase, we set the batch size to $24$ and the initial learning rate to $10^{-5}$.
We train and evaluate our TASSN by using a machine with four GeForce GTX 1080Ti GPUs.

Since end-to-end training a network from scratch with multiple modules is very difficult, we train our TASSN by using a three-stage procedure. 
In the first stage, we train the heatmap estimator with the loss~$\bm{\mathcal{L}_h}$. 
In the second stage, the GCN hand mesh estimator is initialized by using the pre-trained model provided by~\cite{ge2019handshapepose}.
We jointly fine-tune heatmap and hand mesh estimator with the losses $\bm{\mathcal{L}_h}$ and $\bm{\mathcal{L}_m}$ on the target dataset without 3D supervision.
In the final stage, we conduct an end-to-end training for our TASSN and fine-tune the weights of each sub-module.
The model weights of heatmap, GCN hand mesh estimator, and 3D pose estimators are fine-tuned end-to-end.
In this stage, we set $\lambda_s = 0.1$, $\lambda_h = 1$, and $\lambda_c^p = \lambda_c^m = 10$.

\section{Experimental Results}
\begin{table}[t]
\centering
\hspace{-1cm}
\caption{$3$D hand pose estimation results on the STB and MHP datasets. $\uparrow$: higher is better; $\downarrow$: lower is better; The measuring unit of EPE is millimeter (mm).}
    \begin{tabular}{lcccc}
    \hline
&$\text{AUC}_\text{0-50}\uparrow$ &$\text{AUC}_{\text{20-50}}\uparrow$&EPE$\downarrow$\\
\hline
STB Dataset&&&& \\
\hline
TASSN w/o  $\bm{\mathcal{L}}_{c}$&    0.541 &0.735 &24.2\\
TASSN w/o $\bm{\mathcal{L}}_{c}^m$&   0.754   &0.936   &13.6\\
TASSN &                               \bf{0.773} &\bf{0.972}  &\bf{11.3}\\
\hline
\hline
MHP Dataset&&&& \\
\hline
TASSN w/o $\bm{\mathcal{L}}_{c}$        &0.492  &0.677 &28.2\\
TASSN w/o $\bm{\mathcal{L}}_{c}^m$ &0.665  &0.870 &17.5\\
TASSN & \bf{0.689}&\bf{0.892}&\bf{16.2}&\\
\hline
\end{tabular}
\label{table:3Dresult}
\end{table}

\subsection{Ablation Study of Temporal Consistency Losses}

To study the impact of the proposed temporal consistency constraint, we train and evaluate TASSN under the following three settings: 1) TASSN is trained without using temporal consistency loss $\bm{\mathcal{L}}_c$, i.e., without any temporal consistency constraint; 2) TASSN is trained without using temporal consistency loss of hand mesh $\bm{\mathcal{L}}_{c}^m$, i.e., with temporal 3D pose constraint but not 3D mesh constraint; 3) TASSN is trained with all the proposed loss functions.

\begin{figure*}[]
\centering 
\begin{tabular}{cc}
\includegraphics[width=0.42\textwidth]{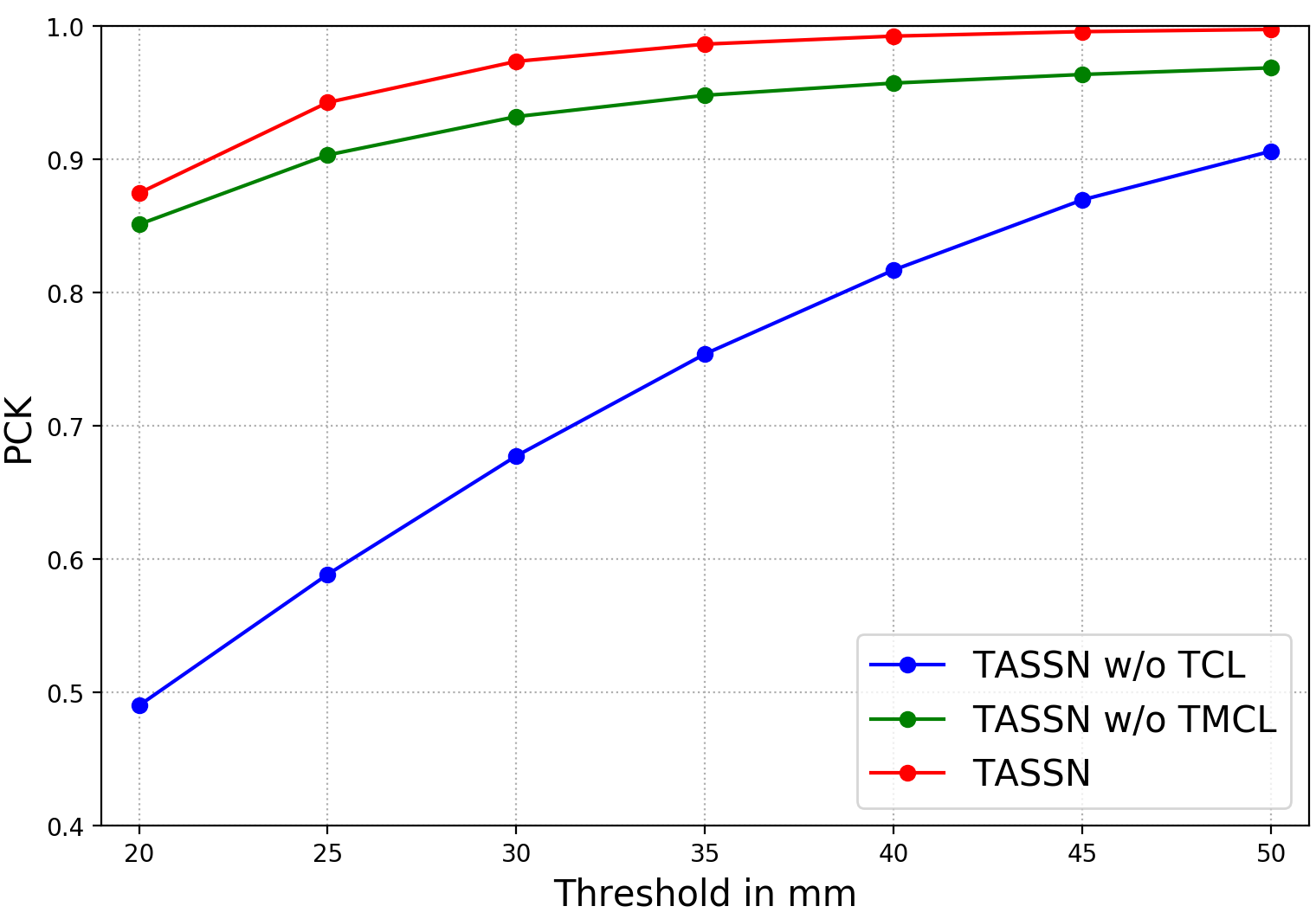}\label{fig:STB}&\quad\quad
\includegraphics[width=0.42\textwidth]{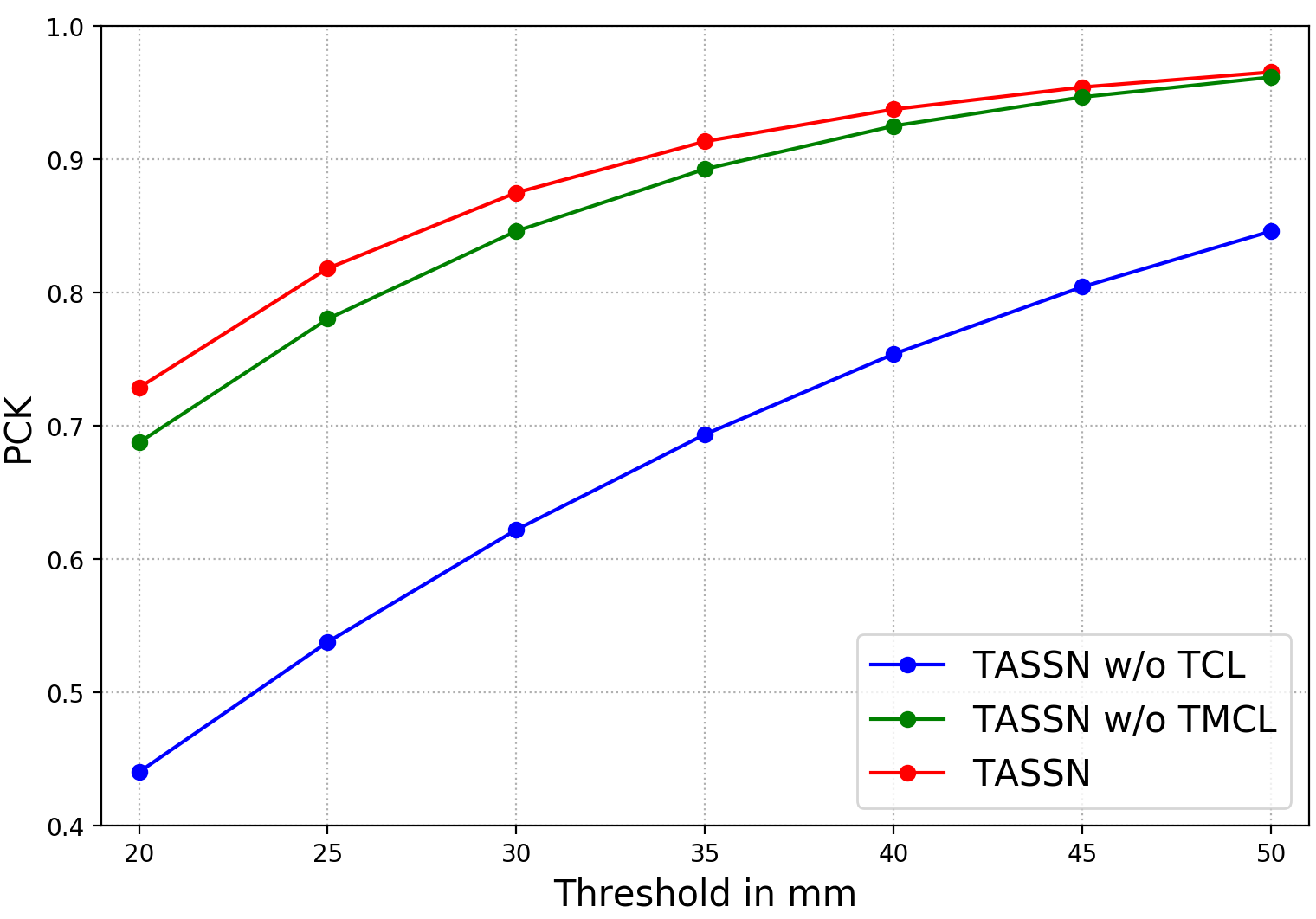}\label{fig:MHP}\\
(a)&\quad(b)\\
\end{tabular}
\caption{Performance in PCK on the (a) STB and (b) MHP datasets. TCL and TMCL denote the losses $\bm{\mathcal{L}}_{c}$ and $\bm{\mathcal{L}}_{c}^m$, respectively.
}
\label{fig:result_setting}
\end{figure*}

Table~\ref{table:3Dresult} shows the evaluation results on two 3D hand pose estimation tasks under the three different settings described above.
The PCK curves corresponding to different settings are shown in Figure~\ref{fig:result_setting}.

\begin{figure*}[]
\centering 
\begin{tabular}{cc}
\includegraphics[width=0.42\textwidth]{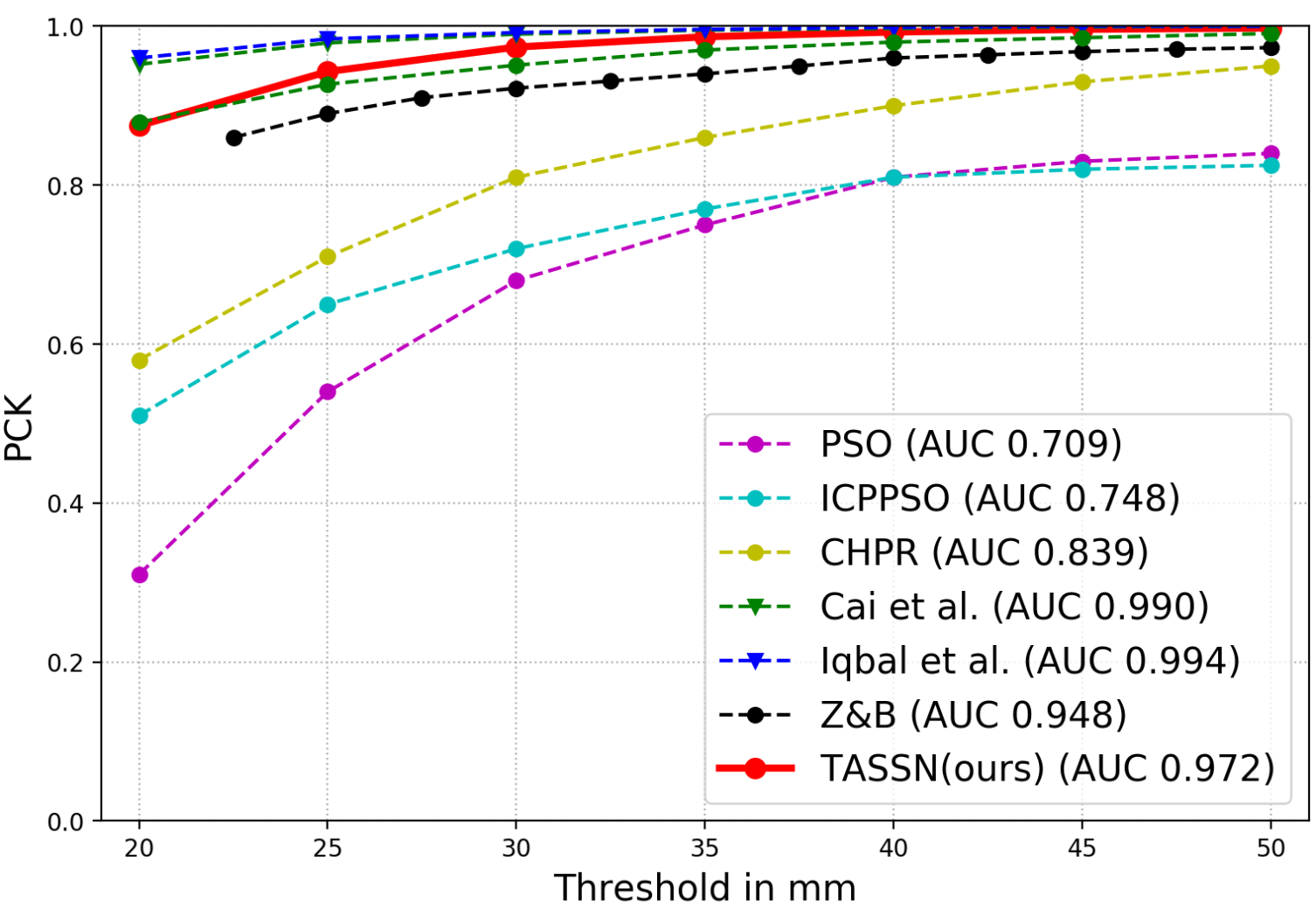}\label{fig:STOA_STB}&\quad\quad
\includegraphics[width=0.42\textwidth]{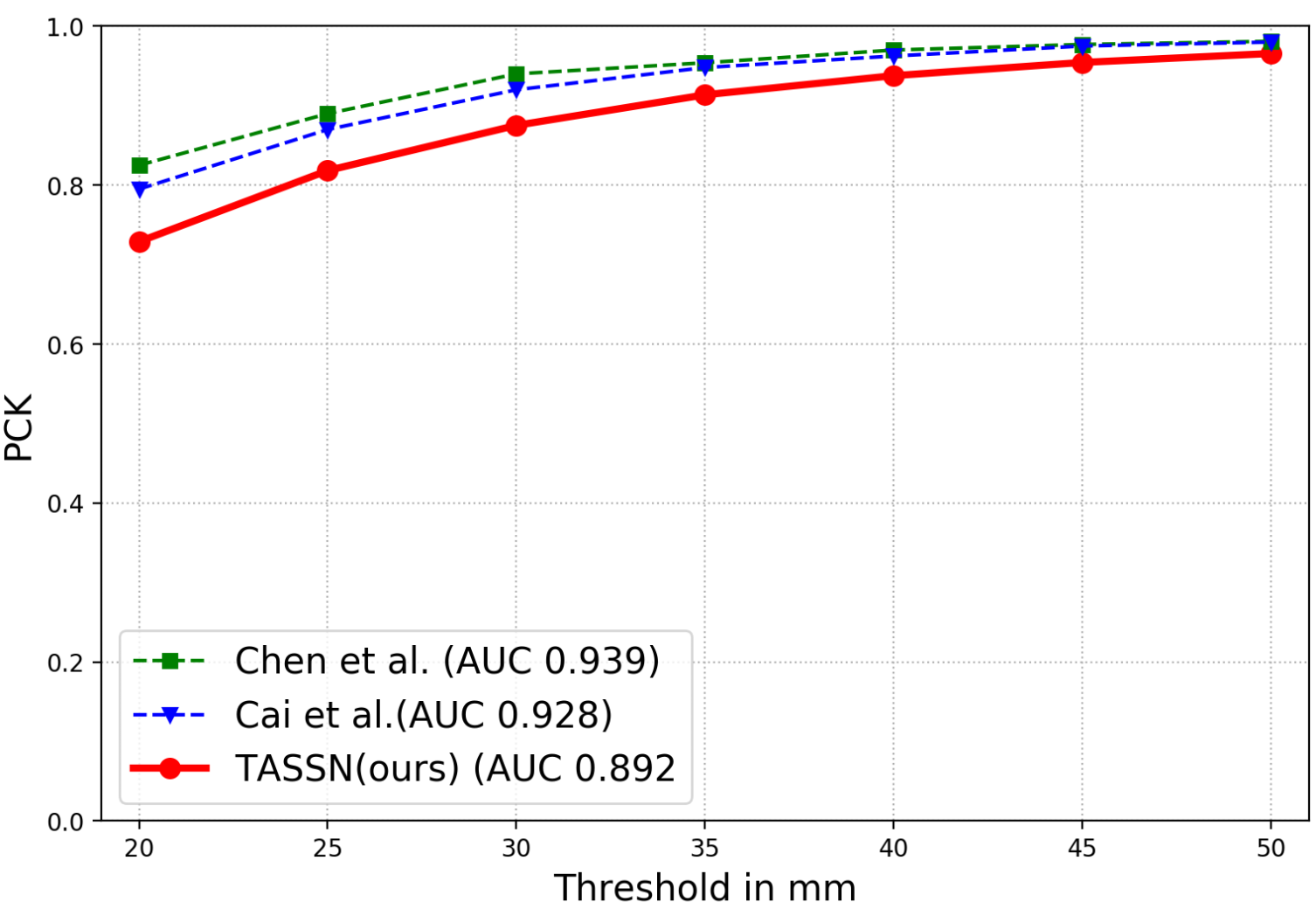}\label{fig:STOA_MHP}\\
(a)&\quad(b)\\
\end{tabular}
\caption{Comparison with the state-of-the-arts. Results in $\text{AUC}_{\text{20-50}}$ on (a) the STB dataset and (b) the MHP dataset.
}
\label{fig:result_stoa}
\end{figure*}

We note the following two observations from the ablation study. First, the temporal consistency constraint is critical for 3D pose estimation accuracy. This is clearly illustrated by comparing the results between settings~$1$ and~$3$.
%
%
As shown in Figure~\ref{fig:result_setting}, TASSN trained with the temporal consistency loss $\bm{\mathcal{L}}_{c}$ (red curve, setting 3) outperforms the TASSN trained without using temporal consistency loss (blue curve, setting 1) by a large margin on both the STB and MHP datasets. 
The quantitative results in Table~\ref{table:3Dresult} show that $\text{AUC}_\text{0-50}$, $\text{AUC}_{20-50}$ and EPE, are improved by 0.232, 0.237, 12.9 on the STB dataset, respectively. 
A similar trend is also observed on the MHP dataset.  
%


Second, imposing temporal mesh consistency constraints is beneficial for 3D pose estimation. This is illustrated by comparing the results between settings~$2$ and setting~$3$.
%
By using the temporal mesh consistency loss $\bm{\mathcal{L}}_c^m$, $\text{AUC}_\text{0-50}$, $\text{AUC}_\text{20-50}$, EPE improves by $0.024$, $0.022$, $1.3$, respectively, on the STB dataset (Table~\ref{table:3Dresult}).
Results on MHP dataset share a same trend: Test AUCs are boosted by including the temporal mesh consistency loss $\bm{\mathcal{L}}_c^m$.
It points out that the temporal mesh consistency loss, as an intermediate constraint, facilitates 3D hand pose estimator learning.

\label{sec:exper}

\begin{figure*}[t]
	\begin{center}
		\begin{tabular}{cccccccccc}
		\includegraphics[width=0.09\textwidth]{SK_164.png}
		\includegraphics[width=0.09\textwidth]{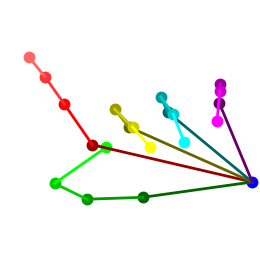}
		\includegraphics[width=0.09\textwidth]{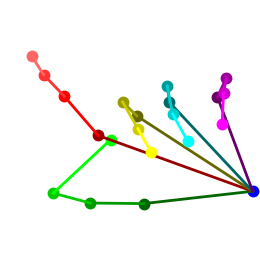}
		\includegraphics[width=0.09\textwidth]{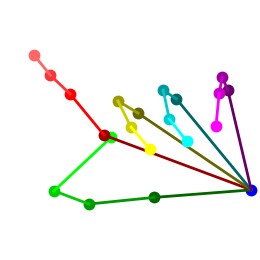}
		\includegraphics[width=0.09\textwidth]{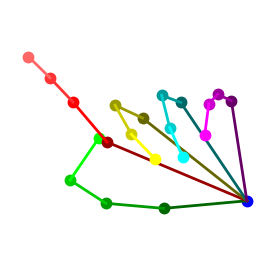}
		\includegraphics[width=0.09\textwidth]{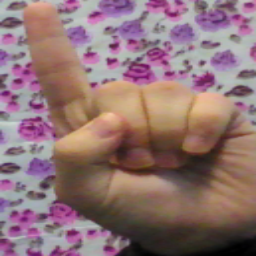}
		\includegraphics[width=0.09\textwidth]{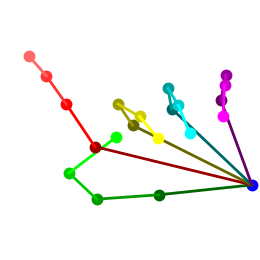}
		\includegraphics[width=0.09\textwidth]{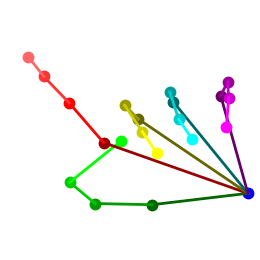}
		\includegraphics[width=0.09\textwidth]{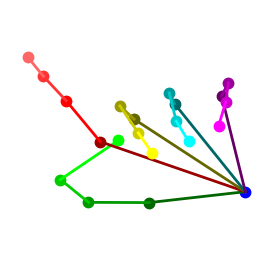}
		\includegraphics[width=0.09\textwidth]{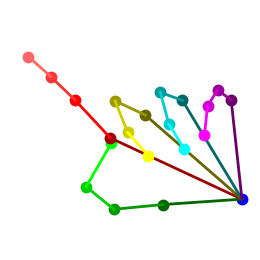}
		\end{tabular}
	\\
	\begin{tabular}{cccccccccc}
		\includegraphics[width=0.09\textwidth]{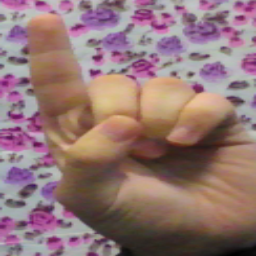}
		\includegraphics[width=0.09\textwidth]{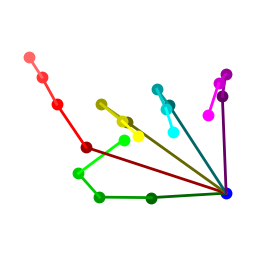}
		\includegraphics[width=0.09\textwidth]{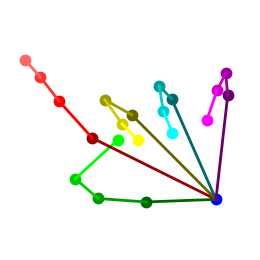}
		\includegraphics[width=0.09\textwidth]{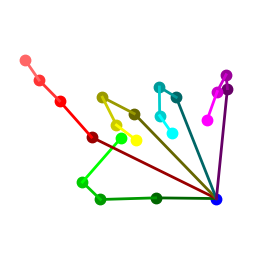}
		\includegraphics[width=0.09\textwidth]{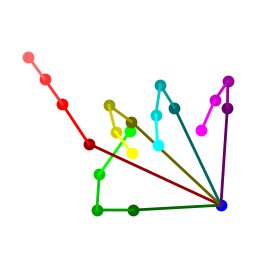}
		\includegraphics[width=0.09\textwidth]{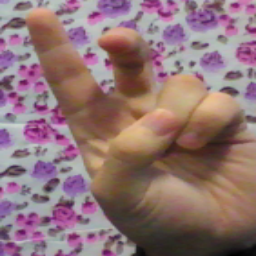}
		\includegraphics[width=0.09\textwidth]{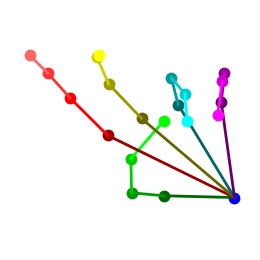}
		\includegraphics[width=0.09\textwidth]{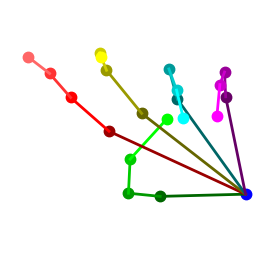}
		\includegraphics[width=0.09\textwidth]{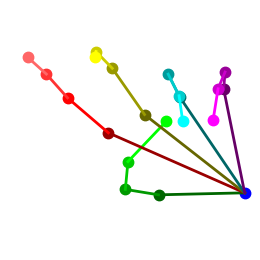}
		\includegraphics[width=0.09\textwidth]{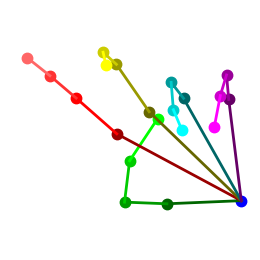}
		\end{tabular}
    \\
    \begin{tabular}{cccccccccc}
		\includegraphics[width=0.09\textwidth]{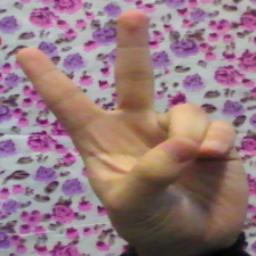}
		\includegraphics[width=0.09\textwidth]{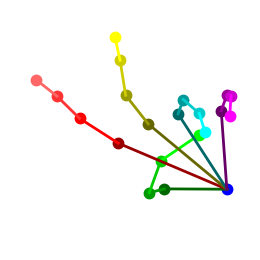}
		\includegraphics[width=0.09\textwidth]{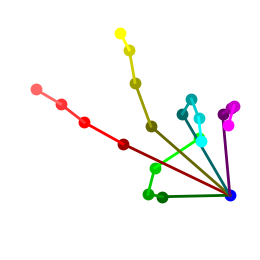}
		\includegraphics[width=0.09\textwidth]{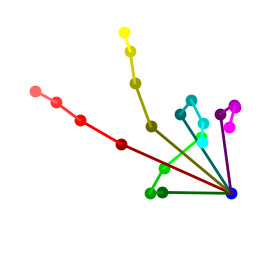}
		\includegraphics[width=0.09\textwidth]{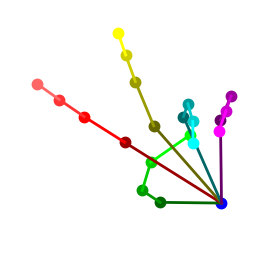}
		\includegraphics[width=0.09\textwidth]{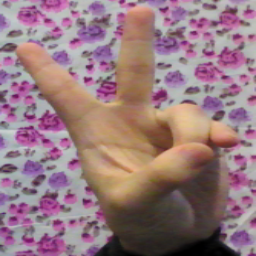}
		\includegraphics[width=0.09\textwidth]{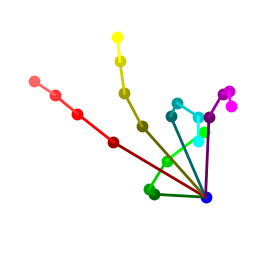}
		\includegraphics[width=0.09\textwidth]{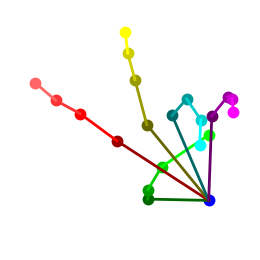}
		\includegraphics[width=0.09\textwidth]{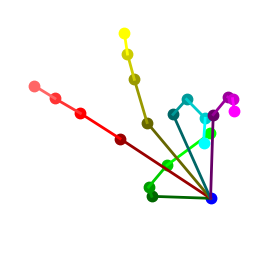}
		\includegraphics[width=0.09\textwidth]{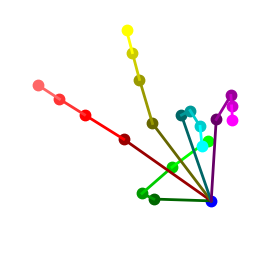}
		\end{tabular}
	    \\
	\end{center}
	\vspace{-.3cm}
	\caption{Comparison among three different settings on the STB dataset. Columns 1 and 6 are RGB images. Columns 2 and 7 are the result by TASSN trained without temporal consistency loss. Columns 3 and 8 are the result by TASSN trained without temporal mesh consistency loss. Columns 4 and 9 are the result by TASSN. Columns 5 and 10 are the ground truth.}
	\label{fig:STBcmp}
\end{figure*}

\begin{figure*}[t]
	\begin{center}
		\begin{tabular}{cccccccccc}
		\includegraphics[width=0.09\textwidth]{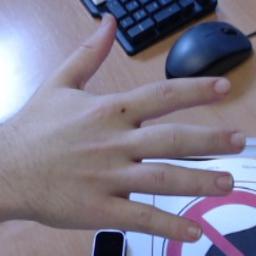}
		\includegraphics[width=0.09\textwidth]{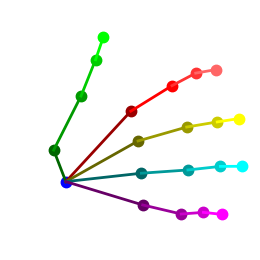}
		\includegraphics[width=0.09\textwidth]{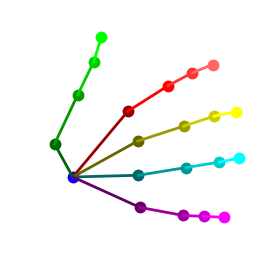}
		\includegraphics[width=0.09\textwidth]{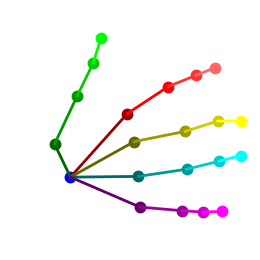}
		\includegraphics[width=0.09\textwidth]{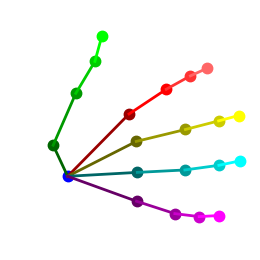}
		\includegraphics[width=0.09\textwidth]{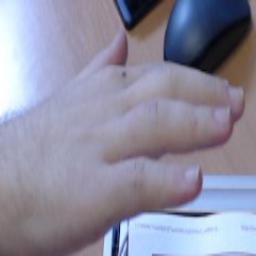}
		\includegraphics[width=0.09\textwidth]{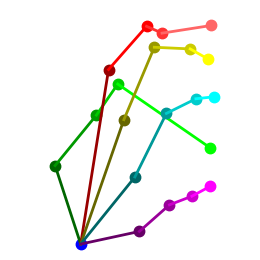}
		\includegraphics[width=0.09\textwidth]{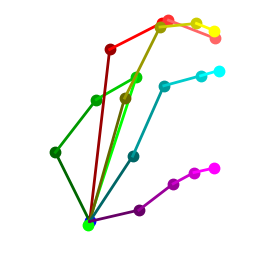}
		\includegraphics[width=0.09\textwidth]{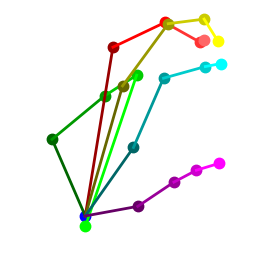}
		\includegraphics[width=0.09\textwidth]{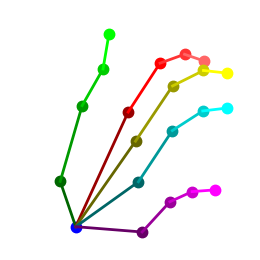}
		\end{tabular}
	\\
	\begin{tabular}{cccccccccc}
		\includegraphics[width=0.09\textwidth]{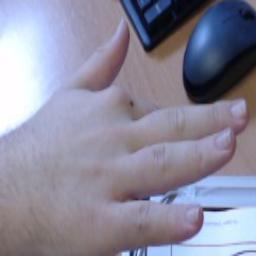}
		\includegraphics[width=0.09\textwidth]{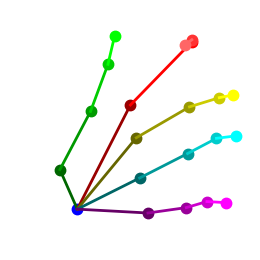}
		\includegraphics[width=0.09\textwidth]{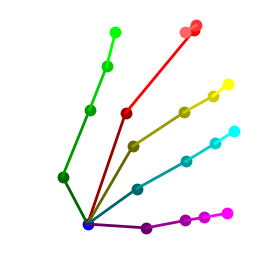}
		\includegraphics[width=0.09\textwidth]{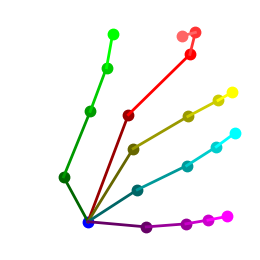}
		\includegraphics[width=0.09\textwidth]{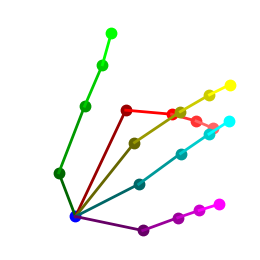}
		\includegraphics[width=0.09\textwidth]{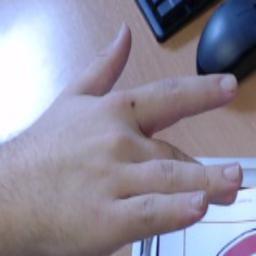}
		\includegraphics[width=0.09\textwidth]{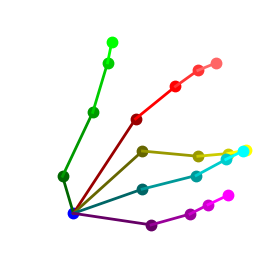}
		\includegraphics[width=0.09\textwidth]{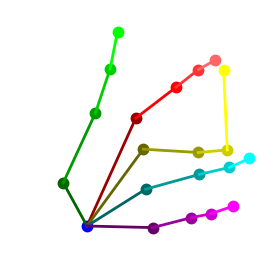}
		\includegraphics[width=0.09\textwidth]{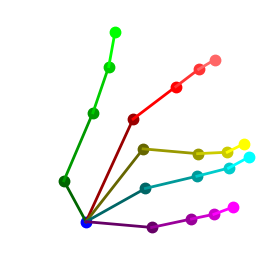}
		\includegraphics[width=0.09\textwidth]{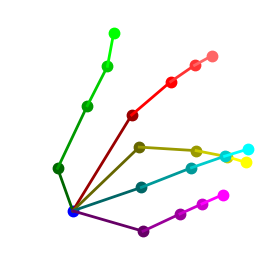}
		\end{tabular}
    \\
    \begin{tabular}{cccccccccc}
		\includegraphics[width=0.09\textwidth]{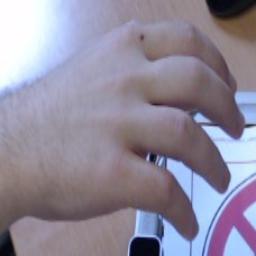}
		\includegraphics[width=0.09\textwidth]{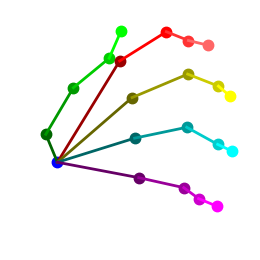}
		\includegraphics[width=0.09\textwidth]{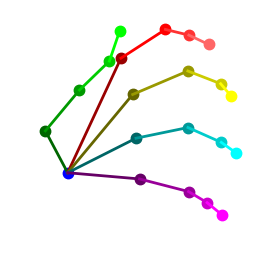}
		\includegraphics[width=0.09\textwidth]{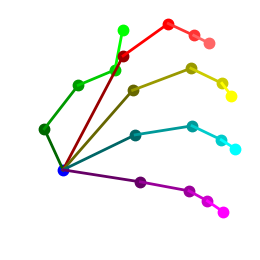}
		\includegraphics[width=0.09\textwidth]{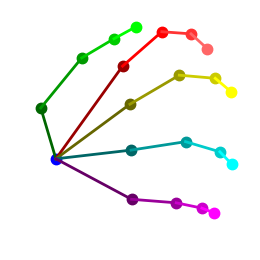}
		\includegraphics[width=0.09\textwidth]{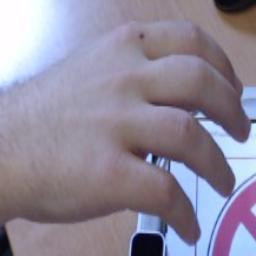}
		\includegraphics[width=0.09\textwidth]{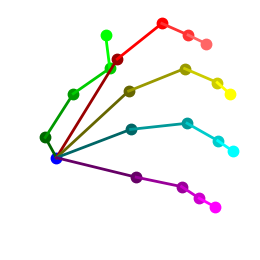}
		\includegraphics[width=0.09\textwidth]{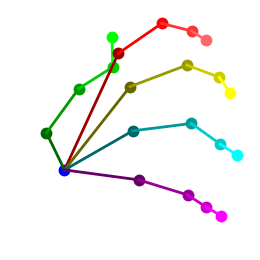}
		\includegraphics[width=0.09\textwidth]{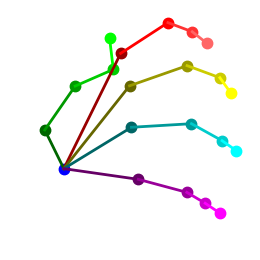}
		\includegraphics[width=0.09\textwidth]{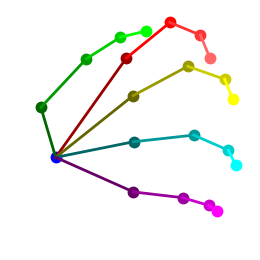}
		\end{tabular}
	    \\
	\end{center}
	\vspace{-.3cm}
	\caption{Comparison among three different settings on the MHP dataset. Columns 1 and 6 are RGB images. Columns 2 and 7 are the result by TASSN trained without temporal consistency loss. Columns 3 and 8 are the result by TASSN trained without temporal mesh consistency loss. Columns 4 and 9 are the result by TASSN. Columns 5 and 10 are the ground truth.
	}
	\vspace{-.3cm}
	\label{fig:MHPcmp}
\end{figure*}

In addition to the quantitative analysis, Figure~\ref{fig:STBcmp} and Figure~\ref{fig:MHPcmp} display some estimated 3D hand poses for visual comparison among these settings on the STB and MHP datasets, respectively. We can see that TASSN, when trained with temporal consistency loss, can produce 3D hand pose estimations highly similar to the ground truth in diverse poses.
It is worth noting that our GCN model is initialized with model~\cite{ge2019handshapepose} pretrained on the STB dataset. Our results on STB demonstrate that the temporal consistency is critical to enforce the 3D constraints, without which 3D prediction accuracy drops substantially (Table~\ref{table:3Dresult}). Moreover, our method generalizes well on other target datasets, e.g., the MHP dataset, where 3D annotations are not used in either model initialization or training. The pose categories and capturing environments are quite different between the two datasets (Figure~\ref{fig:dataset}). The effectiveness of our method on the MHP dataset can only be attributed to the temporal consistency constraint (Figure~\ref{fig:result_setting}).

\subsection{Comparison with the State-of-the-art Methods}

The state-of-the-art methods on both STB and MHP datasets are trained with the 3D annotations, while our method is not. Therefore, we take these methods as the upper bound of our method, and evaluate the performance gaps between these methods and ours.

For the STB dataset, we select six the-state-of-the-art methods for comparison.
The selected methods include PSO~\cite{boukhayma20193d}, ICPPSO~\cite{Chen2018Generating}, CHPR~\cite{zhang20163d}, the method by~Iqbal~\etal~\cite{iqbal2018hand}, Cai~\etal~\cite{cai2018weakly} and the approach by Zimmermann and Brox~\cite{zb2017hand}.
For the MHP dataset, we select two the-state-of-the-art methods for comparison including the approach by~Cai~\etal~\cite{cai2018weakly} and the method by Chen~\etal~\cite{chen2020dggan}.
Figure~\ref{fig:result_stoa}(a) and Figure~\ref{fig:result_stoa}(b) show the comparison results on STB and MHP datasets, respectively.
As expected, TASSN has a performance gap with current state-of-the-art methods on both datasets due to the lack of 3D annotation. However, the performance gaps are relative small. 
In STB dataset, as shown in Figure~\ref{fig:result_stoa}(a), our methods could even beat some of the methods trained with full 3D annotations.

All together, these results illustrate that 3D pose estimator can be trained without using 3D annotations. 
Estimating hand pose and mesh from single frames is challenging due to the ambiguities caused by the missing depth information and high flexibility of joints. These challenges can be partly mitigated by utilizing information from video, in which pose and the mesh are highly constrained by the adjacent frames. Temporal information offers an alternative way of enforcing constraints on 3D models for pose and mesh estimation.  

\section{Conclusions}
We propose a video-based hand pose estimation model, temporal-aware self-supervised network (TASSN), to learn and infer 3D hand pose and mesh from RGB videos.
By leveraging temporal consistency between forward and reverse measurements, TASSN can be trained through self-supervised learning without explicit 3D annotations.
The experimental results show that TASSN achieves reasonably good results with performance comparable to state-of-the-art models trained with 3D ground truth. 

The temporal consistency constraint proposed here offers a convenient and yet effective mechanism for training 3D pose prediction models. Although we illustrate the efficacy of the model without using 3D annotations, it can be used in conjunction with direct supervision with a small number of 3D labeled samples to improve accuracy. 

\paragraph{Acknowledgement.} This work was supported in part by the Ministry of Science and Technology (MOST) under grants MOST 107-2628-E-009-007-MY3, MOST 109-2634-F-007-013, and MOST 109-2221-E-009-113-MY3, and by Qualcomm through a Taiwan University Research Collaboration Project.


\label{sec:conclu}

\clearpage 

{\small
\bibliographystyle{ieee_fullname}
\bibliography{egbib}
}

\end{document}